\documentclass[11pt,a4paper]{article}
\usepackage[hyperref]{acl2020}
\usepackage{times}
\usepackage{latexsym}

\usepackage{graphicx}
\usepackage{caption}
\usepackage{url}
\usepackage{subcaption}
\usepackage{multirow}

\usepackage{microtype}

\aclfinalcopy 

\title{Coach: A Coarse-to-Fine Approach for Cross-domain Slot Filling}

\author{Zihan Liu, Genta Indra Winata, Peng Xu, Pascale Fung \\
Center for Artificial Intelligence Research (CAiRE)\\
Department of Electronic and Computer Engineering\\
The Hong Kong University of Science and Technology, Clear Water Bay, Hong Kong\\
\texttt{zihan.liu@connect.ust.hk}}

\date{}

\begin{document}
\maketitle
\begin{abstract}
As an essential task in task-oriented dialog systems, slot filling requires extensive training data in a certain domain. However, such data are not always available.
Hence, cross-domain slot filling has naturally arisen to cope with this data scarcity problem.
In this paper, we propose a \textbf{Coa}rse-to-fine approa\textbf{ch} (\textbf{Coach}) for cross-domain slot filling. Our model first learns the general pattern of slot entities by detecting whether the tokens are slot entities or not. It then predicts the specific types for the slot entities. In addition, we propose a \textit{template regularization} approach to improve the adaptation robustness by regularizing the representation of utterances based on utterance templates.
Experimental results show that our model significantly outperforms state-of-the-art approaches in slot filling. Furthermore, our model can also be applied to the cross-domain named entity recognition task, and it achieves better adaptation performance than other existing baselines. The code is available at \url{https://github.com/zliucr/coach}.

\end{abstract}

\section{Introduction}
\label{intro}

Slot filling models identify task-related slot types in certain domains for user utterances, and are an indispensable part of task-oriented dialog systems.
Supervised approaches have made great achievements in the slot filling task~\cite{goo2018slot,zhang-etal-2019-joint}, where substantial labeled training samples are needed. However, collecting large numbers of training samples is not only expensive but also time-consuming. To cope with the data scarcity issue, we are motivated to investigate cross-domain slot filling methods, which leverage knowledge learned in the source domains and adapt the models to the target domain with a minimum number of target domain labeled training samples.

\begin{figure}[!ht]
\centering
\begin{subfigure}{.49\textwidth}
    \centering
    \includegraphics[scale=0.76]{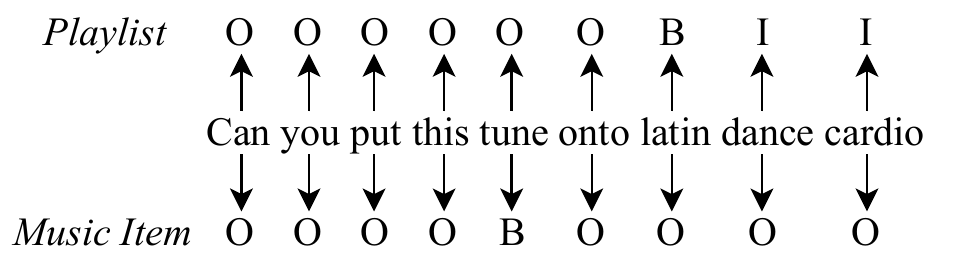}
    \caption{Framework proposed by~\citet{bapna2017towards}.}
    \label{fig:other_framework}
\end{subfigure}
\begin{subfigure}{.49\textwidth}
    \centering
    \includegraphics[scale=0.76]{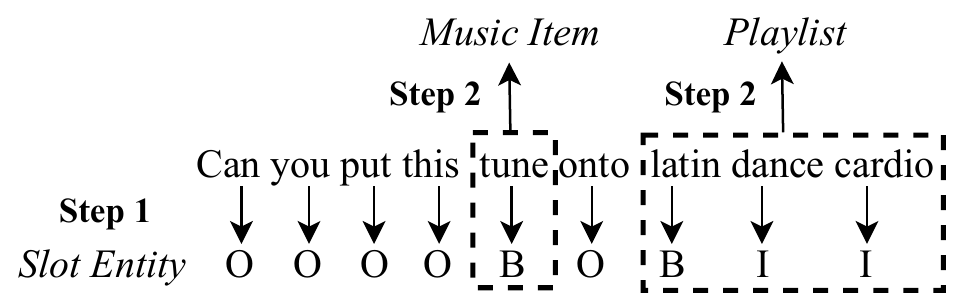}
    \caption{Our proposed framework, \textit{Coach}.}
    \label{fig:our_framework}
\end{subfigure}
\caption{Cross-domain slot filling frameworks.}
\label{fig:frameworks}
\end{figure}

A challenge in cross-domain slot filling is to handle unseen slot types,
which prevents general classification models from adapting to the target domain without any target domain supervision signals.
Recently, \citet{bapna2017towards} proposed a cross-domain slot filling framework, which enables zero-shot adaptation. As illustrated in Figure~\ref{fig:other_framework}, their model conducts slot filling individually for each slot type. It first generates word-level representations, which are then concatenated with the representation of each slot type description, and the predictions are based on the concatenated features for each slot type. 
Due to the inherent variance of slot entities across different domains, it is difficult for this framework to capture the whole slot entity (e.g., ``latin dance cardio'' in Figure~\ref{fig:other_framework}) in the target domain. There also exists a multiple prediction problem. For example, ``tune'' in Figure~\ref{fig:other_framework} could be predicted as ``B'' for both ``music item'' and ``playlist'', which would cause additional trouble for the final prediction.

\begin{figure*}[!ht]
    \centering
    \includegraphics[scale=0.74]{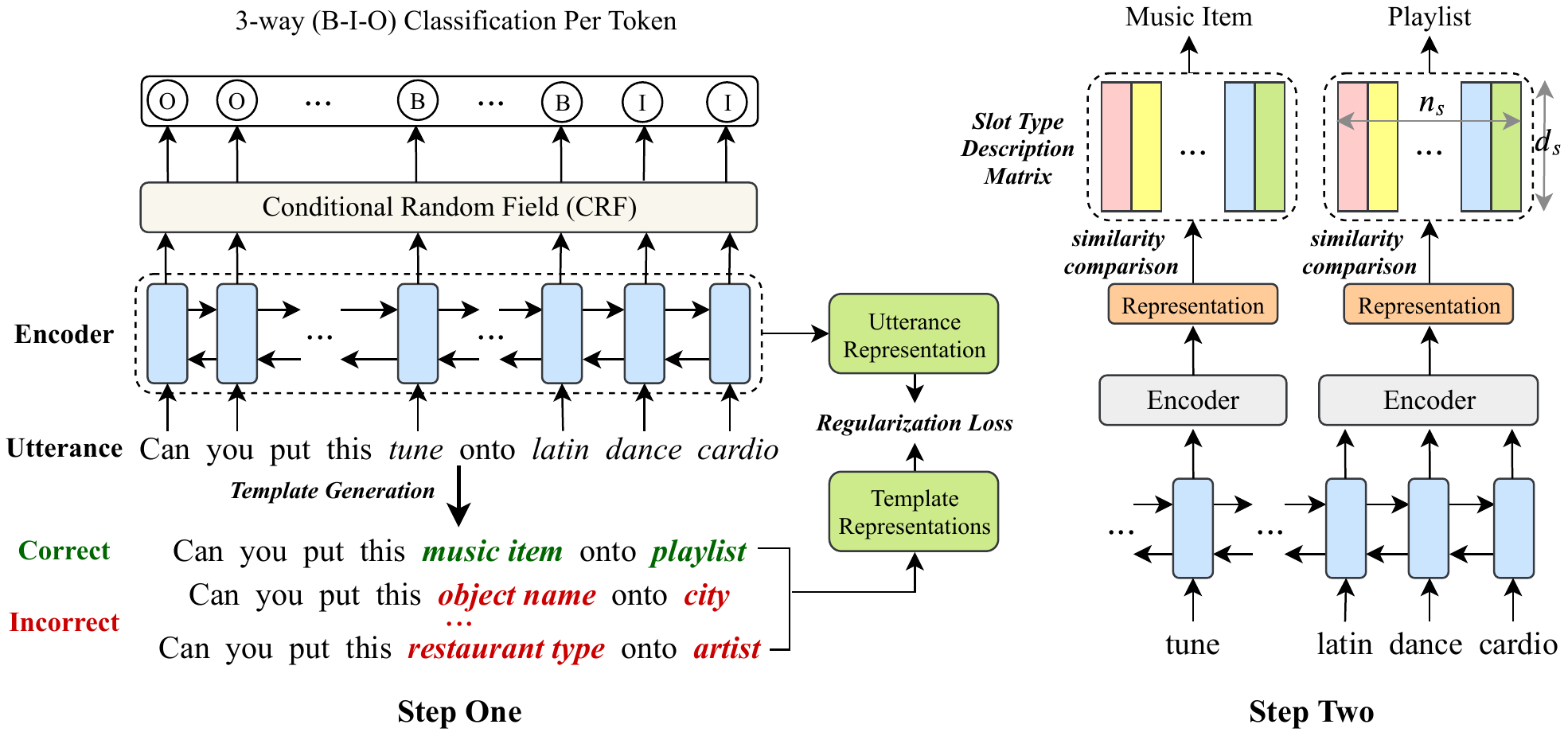}
    \caption{Illustration of our framework, \textit{Coach}, and the template regularization approach.}
    \label{fig:model}
\end{figure*}

We emphasize that in order to capture the whole slot entity, it is pivotal for the model to share its parameters for all slot types in the source domains and learn the general pattern of slot entities. Therefore, as depicted in Figure~\ref{fig:our_framework}, we propose a new cross-domain slot filling framework called \textit{Coach}, a coarse-to-fine approach. It first \textbf{coarsely} learns the slot entity pattern by predicting whether the tokens are slot entities or not. Then, it combines the features for each slot entity and predicts the specific (\textbf{fine}) slot type based on the similarity with the representation of each slot type description. In this way, our framework is able to avoid the multiple predictions problem. 
Additionally, we introduce a \textit{template regularization} method that delexicalizes slot entity tokens in utterances into different slot labels and produces both correct and incorrect utterance templates to regularize the utterance representations. By doing so, the model learns to cluster the representations of semantically similar utterances (i.e., in the same or similar templates) into a similar vector space, which further improves the adaptation robustness.

Experimental results show that our model surpasses the state-of-the-art methods by a large margin in both zero-shot and few-shot scenarios. In addition, further experiments show that our framework can be applied to cross-domain named entity recognition, and achieves better adaptation performance than other existing frameworks.

\section{Related Work}
Coarse-to-fine methods in NLP are best known for syntactic parsing~\cite{charniak2006multilevel,petrov2011coarse}. \citet{zhang2017macro} reduced the search space of semantic parsers by using coarse macro grammars. Different from the previous work, we apply the idea of coarse-to-fine into cross-domain slot filling to handle unseen slot types by separating the slot filling task into two steps~\cite{zhai2017neural,guerini2018toward}.

Coping with low-resource problems where there are zero or few existing training samples has always been an interesting and challenging task~\cite{kingma2014semi,lample2018phrase,liu2019zero,liu2019attention,lin2020xpersona}.
Cross-domain adaptation addresses the data scarcity problem in low-resource target domains~\cite{pan2010cross,jaech2016domain,guo2018multi,jia2019cross,liu2020zero, winata2020learning}.
However, most research studying the cross-domain aspect has not focused on predicting unseen label types in the target domain since both source and target domains have the same label types in the considered tasks~\cite{guo2018multi}. In another line of work, to bypass unseen label types, \citet{ruder2018strong} and \citet{jia2019cross} utilized target domain training samples, so that there was no unseen label type in the target domain.
Recently, based on the framework proposed by~\citet{bapna2017towards} (discussed in Section~\ref{intro}), \citet{lee2019zero} added an attention layer to produce slot-aware representations, and \citet{shah-etal-2019-robust} leveraged slot examples to increase the robustness of cross-domain slot filling adaptation. 

\section{Methodology}
\subsection{Coach Framework}
As depicted in Figure~\ref{fig:model}, the slot filling process in our Coach framework consists of two steps. In the first step, we utilize a BiLSTM-CRF structure~\cite{lample2016neural} to learn the general pattern of slot entities by having our model predict whether tokens are slot entities or not (i.e., 3-way classification for each token).
In the second step, our model further predicts a specific type for each slot entity based on the similarities with the description representations of all possible slot types. To generate representations of slot entities, we leverage another encoder, BiLSTM~\cite{hochreiter1997long}, to encode the hidden states of slot entity tokens and produce representations for each slot entity.

We represent the user utterance with $n$ tokens as $ \textbf{w} = [w_1, w_2, ..., w_{n}]$, and $\textbf{E}$ denotes the embedding layer for utterances. The whole process can be formulated as follows:
\begin{equation}
    [h_1, h_2, ..., h_n] = \textnormal{BiLSTM} (\textbf{E}(\textbf{w})),
\end{equation}
\begin{equation}
    [p_1, p_2, ..., p_n] = \textnormal{CRF}([h_1, h_2, ..., h_n]),
\end{equation}
where 
$[p_1, p_2, ..., p_n]$ are the logits for the 3-way classification. Then, for each slot entity, we take its hidden states to calculate its representation:
\begin{equation}
    r_k = \textnormal{BiLSTM}([h_i, h_{i+1},... h_j]), \label{eq3}
\end{equation}
\begin{equation}
    s_k = M_{desc} \cdot r_k,  \label{eq:measure}
\end{equation}
where $r_k$ denotes the representation of the $k^{th}$ slot entity, $[h_i, h_{i+1}, ..., h_j]$ denotes the BiLSTM hidden states for the $k^{th}$ slot entity, $M_{desc} \in R^{n_s \times d_s} $ is the representation matrix of the slot description ($n_s$ is the number of possible slot types and $d_s$ is the dimension of slot descriptions), and $s_k$ is the specific slot type prediction for this $k^{th}$ slot entity. We obtain the slot description representation $ r^{desc} \in R^{d_s} $ by summing the embeddings of the N slot description tokens (similar to~\citet{shah-etal-2019-robust}):
\begin{equation}
    r^{desc} = \sum_{i=1}^{N} \textbf{E}(t_i),
\end{equation}
where $t_i$ is the $i^{th}$ token and \textbf{E} is the same embedding layer as that for utterances.

\subsection{Template Regularization}
In many cases, similar or the same slot types in the target domain can also be found in the source domains.
Nevertheless, it is still challenging for the model to recognize the slot types in the target domain owing to the variance between the source domains and the target domain. To improve the adaptation ability, we introduce a template regularization method.

As shown in Figure~\ref{fig:model}, we first replace the slot entity tokens in the utterance with different slot labels to generate correct and incorrect utterance templates. Then, we use BiLSTM and an attention layer~\cite{felbo2017using} to generate the utterance and template representations:
\begin{equation}
    e_t = h_t w_a,~~ \alpha_t = \frac{exp(e_t)}{\sum_{j=1}^n exp(e_j)},~~ R=\sum_{t=1}^{n} \alpha_t h_t,
\end{equation}
where $h_t$ is the BiLSTM hidden state in the $t^{th}$ step, $w_a$ is the weight vector in the attention layer and $R$ is the representation for the input utterance or template. 

We minimize the regularization loss functions for the right and wrong templates, which can be formulated as follows:
\begin{equation}
    L^{r} = \textnormal{MSE} (R^{u}, R^{r}),
\end{equation}
\begin{equation}
    L^{w} = -\beta \times \textnormal{MSE} (R^{u}, R^{w}),
\end{equation}
where $R^{u}$ is the representation for the user utterance, $R^{r}$ and $R^{w}$ are the representations of right and wrong templates, we set $\beta$ as one, and \textnormal{MSE} denotes mean square error. Hence, in the training phase, we minimize the distance between $R^{u}$ and $R^{r}$ and maximize the distance between $R^{u}$ and $R^{w}$.
To generate a wrong template, we replace the correct slot entity with another random slot entity, and we generate two wrong templates for each utterance.
To ensure the representations of the templates are meaningful (i.e., similar templates have similar representations) for training $R^{u}$, in the first several epochs, the regularization loss is only to optimize the template representations, and in the following epochs, we optimize both template representations and utterance representations. 


By doing so, the model learns to cluster the representations in the same or similar templates into a similar vector space. Hence, the hidden states of tokens that belong to the same slot type tend to be similar, which boosts the robustness of these slot types in the target domain.



\begin{table*}[!t]
\centering
\resizebox{0.99\textwidth}{!}{
\begin{tabular}{l|cc|cc|cc|cc|cccc}
\hline
Training Setting   & \multicolumn{4}{c|}{Zero-shot} & \multicolumn{4}{c|}{Few-shot on 20 (1\%) samples} & \multicolumn{4}{c}{Few-shot on 50 (2.5\%) samples}         \\ \hline
Domain $\downarrow$ ~Model $\rightarrow$  & CT     & RZT   & Coach  & +TR   & CT        & RZT       & Coach     & +TR      & CT    & \multicolumn{1}{c|}{RZT}   & Coach  & +TR   \\ \hline
AddToPlaylist      & 38.82  & 42.77 & 45.23 & \textbf{50.90} & 58.36     & \textbf{63.18}     & 58.29    & 62.76    & 68.69 & \multicolumn{1}{c|}{\textbf{74.89}} & 71.63 & 74.68 \\
BookRestaurant     & 27.54  & 30.68 & 33.45 & \textbf{34.01} & 45.65     & 50.54     & 61.08    & \textbf{65.97}    & 54.22 & \multicolumn{1}{c|}{54.49} & 72.19 & \textbf{74.82} \\
GetWeather         & 46.45  & 50.28 & 47.93 & \textbf{50.47} & 54.22     & 58.86     & 67.61    & \textbf{67.89}    & 63.23 & \multicolumn{1}{c|}{58.87} & \textbf{81.55} & 79.64 \\
PlayMusic          & 32.86  & \textbf{33.12} & 28.89 & 32.01 & 46.35     & 47.20     & 53.82    & \textbf{54.04}    & 54.32 & \multicolumn{1}{c|}{59.20} & 62.41 & \textbf{66.38} \\
RateBook           & 14.54  & 16.43 & \textbf{25.67} & 22.06 & 64.37     & 63.33     & \textbf{74.87}    & 74.68    & 76.45 & \multicolumn{1}{c|}{76.87} & \textbf{86.88} & 84.62 \\
SearchCreativeWork & 39.79  & 44.45 & 43.91 & \textbf{46.65} & 57.83     & \textbf{63.39}     & 60.32    & 57.19    & 66.38 & \multicolumn{1}{c|}{\textbf{67.81}} & 65.38 & 64.56 \\
FindScreeningEvent & 13.83  & 12.25 & \textbf{25.64} & 25.63 & 48.59     & 49.18     & 66.18    & \textbf{67.38}    & 70.67 & \multicolumn{1}{c|}{74.58} & 78.10 & \textbf{83.85} \\ \hline
Average F1            & 30.55  & 32.85 & 35.82 & \textbf{37.39} & 53.62     & 56.53     & 63.17    & \textbf{64.27}    & 64.85 & \multicolumn{1}{c|}{66.67} & 74.02 & \textbf{75.51} \\ \hline
\end{tabular}
}
\caption{Slot F1-scores based on standard BIO structure for SNIPS. Scores in each row represents the performance of the leftmost target domain, and TR denotes template regularization.
}
\label{table:slotfilling}
\end{table*}

\section{Experiments}
\subsection{Dataset}
We evaluate our framework on SNIPS~\cite{coucke2018snips}, a public spoken language understanding dataset which contains 39 slot types across seven domains (intents) and $\sim$2000 training samples per domain. To test our framework, each time, we choose one domain as the target domain and the other six domains as the source domains.

Moreover, we also study another adaptation case
where there is no unseen label in the target domain. We utilize the CoNLL-2003 English named entity recognition (NER) dataset as the source domain~\cite{tjong-kim-sang-de-meulder-2003-introduction}, and the CBS SciTech News NER dataset from~\citet{jia2019cross} as the target domain. These two datasets have the same four types of entities, namely, PER (person), LOC (location), ORG (organization), and MISC (miscellaneous).

\subsection{Baselines}
We use word-level~\cite{bojanowski2017enriching} and character-level~\cite{hashimoto2017joint} embeddings for our model as well as all the following baselines for a fair comparison.
\paragraph{Concept Tagger (CT)} \citet{bapna2017towards} proposed a slot filling framework that utilizes slot descriptions to cope with the unseen slot types in the target domain.

\paragraph{Robust Zero-shot Tagger (RZT)} Based on CT, \citet{shah-etal-2019-robust} leveraged example values of slots to improve robustness of cross-domain adaptation.



\paragraph{BiLSTM-CRF} This baseline is only for the cross-domain NER. Since there is no unseen label in the NER target domain, the BiLSTM-CRF~\cite{lample2016neural} uses the same label set for the source and target domains and casts it as an entity classification task for each token, which is applicable in both zero-shot and few-shot scenarios.

\subsection{Training Details}
We use a 2-layer BiLSTM with a hidden size of 200 and dropout rate of 0.3 for both the template encoder and utterance encoder. Note that the parameters in these two encoders are not shared.
The BiLSTM for encoding the hidden states of slot entity tokens has one layer with a hidden size of 200, which would output the same dimension as the concatenated word-level and char-level embeddings. We use Adam optimizer with a learning rate of 0.0005. Cross-entropy loss is leveraged to train the 3-way classification in the first step, and the specific slot type predictions are used in the second step. We split 500 data samples in the target domain as the validation set for choosing the best model and the remainder are used for the test set.
We implement the model in CT and RZT and follow the same setting as for our model for a fair comparison.

\section{Results \& Discussion}

\subsection{Cross-domain Slot Filling}
\paragraph{Quantitative Analysis}
As illustrated in Table~\ref{table:slotfilling}, we can clearly see that our models are able to achieve significantly better performance than the current state-of-the-art approach (RZT).
The CT framework suffers from the difficulty of capturing the whole slot entity, while our framework is able to recognize the slot entity tokens by sharing its parameters across all slot types. Based on the CT framework, the performance of RZT is still limited, and Coach outperforms RZT by a $\sim$3\% F1-score in the zero-shot setting. Additionally, template regularization further improves the adaptation robustness by helping the model cluster the utterance representations into a similar vector space based on their corresponding template representations.

Interestingly, our models achieve impressive performance in the few-shot scenario. In terms of the averaged performance, our best model (Coach+TR) outperforms RZT by $\sim$8\% and $\sim$9\% F1-scores on the 20-shot and 50-shot settings, respectively.
We conjecture that our model is able to better recognize the whole slot entity in the target domain and map the representation of the slot entity belonging to the same slot type into a similar vector space to the representation of this slot type based on Eq (\ref{eq:measure}). This enables the model to quickly adapt to the target domain slots.


\begin{table}[]
\centering
\resizebox{0.48\textwidth}{!}{
\begin{tabular}{l|cc|cc|cc}
\hline
\multirow{2}{*}{\begin{tabular}[c]{@{}l@{}}Target \\ Samples$^\ddagger$\end{tabular}} & \multicolumn{2}{c|}{0 samples} & \multicolumn{2}{c|}{20 samples} & \multicolumn{2}{c}{50 samples} \\ \cline{2-7} & unseen         & seen          & unseen          & seen          & unseen         & seen          \\ \hline
CT        & 27.1           & 44.18         & 50.13           & 61.21         & 62.05          & 69.64         \\
RZT        & 28.28          & 47.15         & 52.56           & 63.26         & 63.96          & 73.10          \\ \hline
Coach      & 32.89          & 50.78         & 61.96           & 73.78         & 74.65          & 76.95         \\
Coach+TR    & \textbf{34.09}          & \textbf{51.93}         & \textbf{64.16}           & \textbf{73.85}         & \textbf{76.49}          & \textbf{80.16}         \\ \hline
\end{tabular}
}
\caption{Averaged F1-scores for seen and unseen slots over all target domains. $^\ddagger$ represent the number of training samples utilized for the target domain.}
\label{table:seen_unseen}
\end{table}

\paragraph{Analysis on Seen and Unseen Slots}
We take a further step to test the models on seen and unseen slots in target domains to analyze the effectiveness of our approaches. To test the performance, we split the test set into ``unseen'' and ``seen'' parts. An utterance is categorized into the ``unseen'' part as long as there is an unseen slot (i.e., the slot does not exist in the remaining six source domains) in it. Otherwise we categorize it into the ``seen'' part. The results for the ``seen'' and ``unseen'' categories are shown in Table~\ref{table:seen_unseen}.
We observe that our approaches generally improve on both unseen and seen slot types compared to the baseline models. For the improvements in the unseen slots, our models are better able to capture the unseen slots since they explicitly learn the general pattern of slot entities.
Interestingly, our models also bring large improvements in the seen slot types. We conjecture that it is also challenging to adapt models to seen slots due to the large variance between the source and target domains. For example, slot entities belonging to the ``object type'' in the ``RateBook'' domain are different from those in the ``SearchCreativeWork'' domain. Hence, the baseline models might fail to recognize these seen slots in the target domain, while our approaches can adapt to the seen slot types more quickly in comparison.
In addition, we observe that template regularization improves performance in both seen and unseen slots, which illustrates that clustering representations based on templates can boost the adaptation ability.

\subsection{Cross-domain NER}
From Table~\ref{table:ner}, we see that the Coach framework is also suitable for the case where there are no unseen labels in the target domain in both the zero-shot and few-shot scenarios, while CT and RZT are not as effective as BiLSTM-CRF. 
However, we observe that template regularization loses its effectiveness in this task, since the text in NER is relatively more open, which makes it hard to capture the templates for each label type.

\begin{table}[]
\centering
\resizebox{0.35\textwidth}{!}{
\begin{tabular}{l|cc}
\hline
Target Samples    & 0     & 50    \\ \hline
CT~(\citet{bapna2017towards})  & 61.43 & 65.85 \\
RZT~(\citet{shah-etal-2019-robust})   & 61.94    & 65.21     \\ 
BiLSTM-CRF        & 61.77    & 66.57     \\ \hline
Coach & 64.08   & \textbf{68.35}     \\
Coach + TR   & \textbf{64.54}   & 67.45     \\ \hline
\end{tabular}
}
\caption{F1-scores on the NER target domain (CBS SciTech News).}
\label{table:ner}
\end{table}

\begin{table}[]
\centering
\resizebox{0.49\textwidth}{!}{
\begin{tabular}{l|ccc|ccc}
\hline
\multirow{2}{*}{Task} & \multicolumn{3}{c|}{zero-shot} & \multicolumn{3}{c}{few-shot on 50 samples} \\ \cline{2-7} 
                      & sum      & trs      & bilstm     & sum          & trs          & bilstm         \\ \hline
Slot Filling          & 33.89    & 34.33    & \textbf{35.82}    & 73.8         & 72.66        & \textbf{74.02}        \\ \hline
NER                   & 63.04    & 63.29    & \textbf{64.47}    & 66.98        & 68.04        & \textbf{68.35}        \\ \hline
\end{tabular}
}
\caption{Ablation study in terms of the methods to encode the entity tokens on Coach.}
\label{table:ablation}
\end{table}

\subsection{Ablation Study}
We conduct an ablation study in terms of the methods to encode the entity tokens (described in Eq. (\ref{eq3})) to investigate how they affect the performance. Instead of using BiLSTM, we try two alternatives. One is to use the encoder of Transformer (trs)~\cite{vaswani2017attention}, and the other is to simply sum the hidden states of slot entity tokens. From Table~\ref{table:ablation}, we can see that there is no significant performance difference among different methods, and we observe that using BiLSTM to encode the entity tokens generally achieves better results.

\section{Conclusion}
We introduce a new cross-domain slot filling framework to handle the unseen slot type issue. Our model shares its parameters across all slot types and learns to predict whether input tokens are slot entities or not. Then, it detects concrete slot types for these slot entity tokens based on the slot type descriptions. 
Moreover, template regularization is proposed to improve the adaptation robustness further.
Experiments show that our model significantly outperforms existing cross-domain slot filling approaches, and it also achieves better performance for the cross-domain NER task, where there is no unseen label type in the target domain.

\section*{Acknowledgments}
This work is partially funded by ITF/319/16FP and MRP/055/18 of the Innovation Technology Commission, the Hong Kong SAR Government.

\bibliography{acl2020}
\bibliographystyle{acl_natbib}

\end{document}